%% file: ms.tex
\pdfoutput=1
\RequirePackage{amsmath}         %

\documentclass[letterpaper, 10pt, conference]{ieeeconf}      %

\IEEEoverridecommandlockouts                              %

\let\labelindent\relax                                    %

\usepackage{graphicx}            %
\usepackage{amssymb}             %
\usepackage{amsfonts}            %
\usepackage{enumitem}            %
\usepackage{multirow}            %
\usepackage{siunitx}             %
\usepackage{url}                 %
\usepackage{xspace}              %
\usepackage[T1]{fontenc}         %
\usepackage[hidelinks]{hyperref} %
\usepackage{wrapfig}
\usepackage{todonotes}
\usepackage{units}
\usepackage{eurosym}
\usepackage{balance}

\graphicspath{{images/}}
\DeclareGraphicsExtensions{.pdf,.png,.jpg,.jpeg}

\newcommand{\degreem}{^{\circ}} %

\newcommand{\seclabel}[1]{\label{sec:#1}}

\newcommand{\figlabel}[1]{\label{fig:#1}}
\newcommand{\tablabel}[1]{\label{tab:#1}}

\newcommand{\figref}[1]{Fig.~\ref{fig:#1}\xspace}
\newcommand{\tabref}[1]{Table~\ref{tab:#1}\xspace}

\newcommand{\nop}{NimbRo\protect\nobreakdash-OP\xspace}
\newcommand{\noptwo}{NimbRo\protect\nobreakdash-OP2\xspace}
\newcommand{\nopx}{NimbRo\protect\nobreakdash-OP2X\xspace}

\newcommand{\degree}{$\degreem$\xspace}

\title{\LARGE \textbf{NimbRo-OP2X: Adult-sized Open-source 3D Printed Humanoid Robot}}

\author{Grzegorz Ficht, Hafez Farazi, Andr\'{e} Brandenburger, Diego Rodriguez, \\Dmytro Pavlichenko, Philipp Allgeuer, Mojtaba Hosseini, and Sven Behnke%
\thanks{All authors are with the Autonomous Intelligent Systems (AIS) Group, Computer Science Institute VI,
        University of Bonn, Germany. Email: {\tt\small ficht@ais.uni-bonn.de}. This work was partially
        funded by grant BE 2556/13 of the German Research Foundation (DFG).}}

\usepackage{eso-pic}

\AtBeginDocument{\AddToShipoutPictureFG*{\AtTextUpperLeft{\put(0,\LenToUnit{9pt}){\parbox{\textwidth}{\centering\bfseries
International Conference on Humanoid Robots (Humanoids), Beijing, China, 2018
}}}}}
\begin{document}
\maketitle
\thispagestyle{empty}
\pagestyle{empty}

\begin{abstract}

Humanoid robotics research depends on capable robot platforms, but
recently developed advanced platforms are often not available to other research groups, expensive, dangerous to operate, or closed-source. 
The lack of available platforms forces researchers to work with smaller robots, which have less strict
dynamic constraints or with simulations, which lack many real-world effects. 
We developed \nopx to address this need.
At a height of 135\,cm our robot is large enough to interact in a human environment. Its low 
weight of only 19\,kg makes the operation of the robot safe and easy, as no special operational 
equipment is necessary. Our robot is equipped with a fast onboard computer and a GPU to accelerate parallel computations. We extend our already open-source software by a deep-learning based vision system and gait parameter optimisation.
The \nopx was evaluated during RoboCup 2018 in Montre\'al, Canada, where it won all possible awards in the Humanoid AdultSize class.

\end{abstract}

\input{introduction.tex}
\input{related_work.tex}
\input{hardware.tex} %
\input{software.tex} %

\input{evaluation.tex}%

\input{conclusions.tex}

\section{Acknowledgements}
\footnotesize
This work was partially funded by grant BE 2556/13 of the German Research Foundation (DFG).

\bibliographystyle{IEEEtran}
\balance
\bibliography{ms}

\end{document}

%% file: introduction.tex
\section{Introduction}

Using humanoid robots in daily life scenarios has been the motivation for 
many research and development efforts. More than 40 years ago, the Waseda robot~\cite{wabot} 
started the quest for building humanoid robots that could coexist in our environment.
Since then, much work has been done in this direction, but we still do not have any humanoid robot 
capable of coping with the uncertainty and complexity of real-life scenarios. From a technical point of view, the 
difficulties arise not only from the lack of actuation capacity for the generation of dexterous movements to interact with the environment, 
but also from deficits on the cognitive level to understand the surrounding and plan tasks accordingly. 

Humanoid robotics has recently seen progress in terms of hardware, with the introduction of multiple new and capable platforms.
The technological advancements made by these platforms however, are met with drawbacks. The precision technology used and their complex design, both require 
that the robot is operated by an experienced user with safety equipment such as gantries to minimise possible risk. 
Many of these advanced platforms are not sold to other researchers. If they were offered, their price tag would be stiff.
Due to the closed-source nature of such projects, only a selected few have
access to these platforms. This forces other research groups to work with smaller robots in scaled-down environments. 
Due to the scale difference, the results may not always translate into real-world applications.
Another option is to use simulations, which may lack faithful reproduction of real-world effects. 

In this paper, we introduce \nopx, which is a \SI{135}{cm} tall, humanoid robot with a completely 3D printed structure and
deep learning capabilities. \nopx is a substantial improvement over our previous \nop and \noptwo robots.
By incorporating new intelligent actuators and a GPU-enabled computing unit into the design, we make  \nopx
applicable in more general use cases. 
In contrast to similarly sized platforms~(\figref{fig_teaser}), \nopx is very light and easy to operate, while having considerable 
torque output to efficiently walk and perform full-body motions. 
By using widely available manufacturing techniques 
and open sourcing both, the hardware and software, we aim to foster accelerated humanoid research in real-world applications. 
\begin{figure}[!t]
\includegraphics[width=1\linewidth]{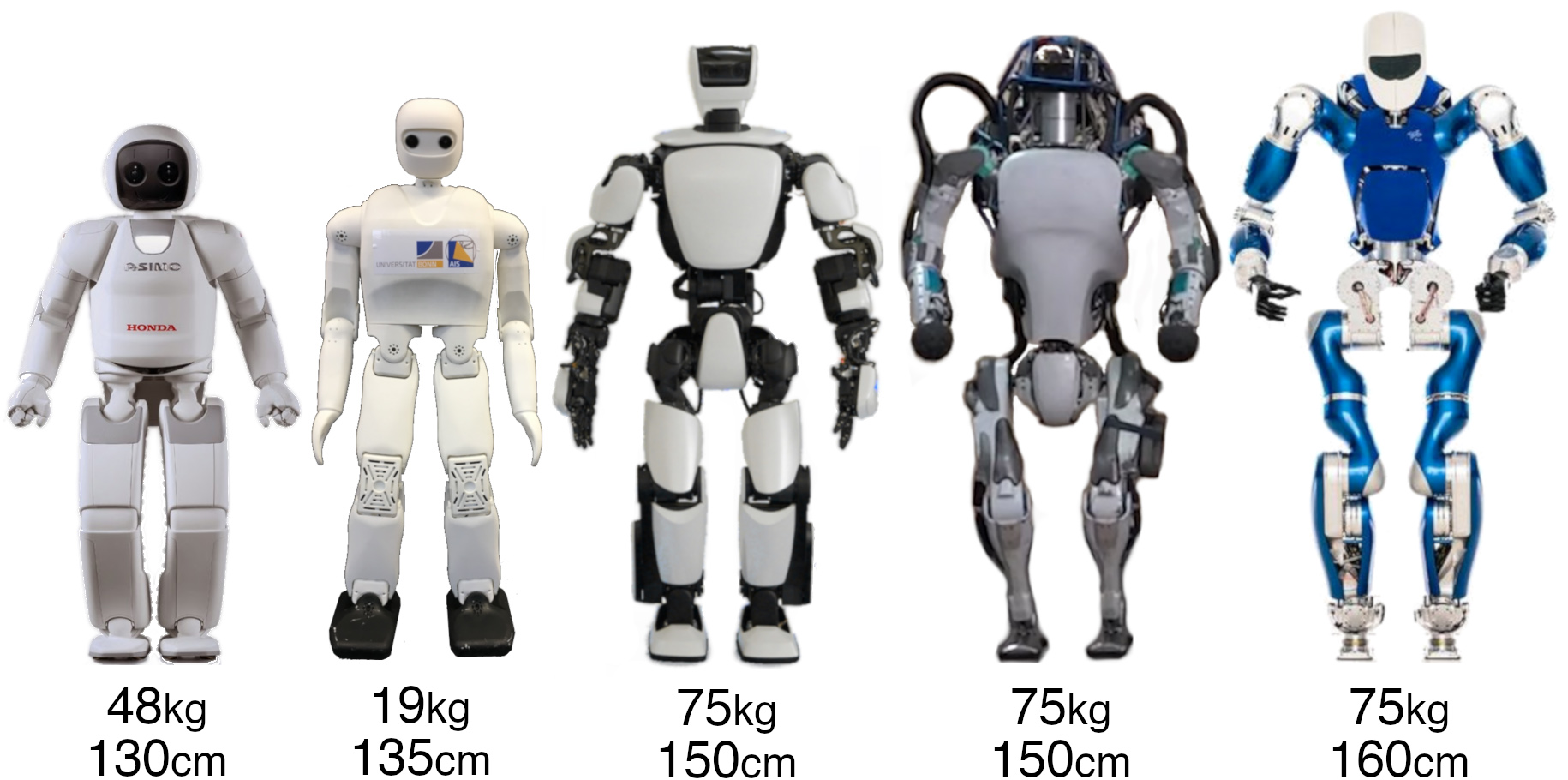}
\caption{\nopx in comparison to other platforms, in terms of height and weight. From left to right: Honda Asimo, \nopx, Toyota T-HR3, Boston Dynamics Atlas, DLR-TORO. 
Out of these platforms, only the \nopx is openly available to any research group.}
\figlabel{fig_teaser}\vspace{-3ex}
\end{figure}

%% file: related_work.tex
\section{Related Work}

For several years, research on motion generation for humanoid robots was carried out mainly by a few companies or universities 
because of the large costs associated with platform development. The most prominent examples include the HRP2~\cite{hrp2} and Honda's Asimo~\cite{asimo}.
The situation did not change only until recently, when the rising interest in humanoid robotics led to the emergence of new human-sized platforms.
Robots such as Petman~\cite{nelson2012petman}, DLR-TORO~\cite{englsberger2014overview}, NASA Valkyrie~\cite{radford2015valkyrie} or the 
recently introduced Toyota T-HR3~\cite{thr3} are still very expensive or, more importantly, these robots  \textit{cannot} be purchased by other research groups.
In 2017, we introduced a larger open-source robot~\cite{ficht2017nop2}, that in comparison with other platforms
could be considered as low-cost thanks to its 3D-printed structure and commercial off-the-shelf actuators.
The robot proved to be capable enough to win the RoboCup 2017 soccer tournament in its debut~\cite{Ficht2018Grown}. 

Shortcutting the limiting factor of hardware availability, advancing the state of the art can be done by means of simulation~\cite{farchy}.
However, simulators still have several challenges to face until they will provide comparable results to reality.
On the other hand, simulations can be performed without any cost, risk, and---most importantly---fast.
In our recent work~\cite{Rodriguez2018Combining}, simulation and real-robot experiments were combined for gait optimisation of smaller robots.
In this paper, we extend this work to our new larger robot, with more complex gait sequences, to achieve
a self-tuned gait that is able to withstand strong impacts. 

Deep learning currently plays a key role in enhancing the cognitive abilities of robots. It has made its way to other research areas in robotics such as: speech recognition~\cite{hinton2012deep}, motion planning~\cite{finn2017deep}, 
learning dynamics~\cite{mnih2015human}, and others. Semantic segmentation addresses most of the perception needs for autonomous robots. Due to heavy computational requirements, many deep learning based approaches for semantic segmentation are not suitable for robotics applications. Romera at al.~\cite{romera2018erfnet} recently proposed "ERFNet" which is ideal for lightweight computers like Jetson TX1. ERFNet is an encoder-decoder architecture for real-time semantic segmentation tasks. We hope that open-sourcing an AdultSize low-cost humanoid platform, 
equipped with a lightweight GPU for parallel computing, will foster research not only limited to motion planning but deep learning as well.

%% file: hardware.tex
\section{Hardware Design}
\seclabel{hardware_design}

\begin{table}
\renewcommand{\arraystretch}{1.2}
\caption{\nopx specifications.}
\tablabel{OP2X_specs}
\centering
\footnotesize
\begin{tabular}{c c c}
\hline
\textbf{Type} & \textbf{Specification} & \textbf{Value}\\
\hline
\multirow{4}{*}{\textbf{General}} & Height \& Weight & \SI{135}{cm}, \SI{19}{kg}\\
& Battery & 4-cell LiPo (\SI{14.8}{V}, \SI{8.0}{Ah})\\
& Battery life & \SI{20}{}--\SI{40}{\minute}\\
& Material & Polyamide 12 (PA12)\\
\hline
\multirow{6}{*}{\textbf{PC}}
& Mainboard & Z370 Chipset, Mini-ITX\\
& CPU & Intel Core i7-8700T, \SI{2.7}{}--\SI{4.0}{GHz}\\
& GPU & GTX 1050 Ti, 768 CUDA Cores\\
& Memory & \SI{4}{GB} DDR4 RAM, \SI{120}{GB} SSD\\
& Network & Ethernet, Wi-Fi, Bluetooth\\
& Other & 8$\,\times\,$USB 3.1, 2$\,\times\,$HDMI, DisplayPort\\
\hline
\multirow{4}{*}{\textbf{Actuators}} 
& Total & 34$\,\times\,$Robotis XM-540-W270-R\\
& Stall torque & \SI{12.9}{Nm} \\
& No load speed & \SI{37}{rpm} \\
& Control mode & Torque, Velocity, Position, Multi-turn\\
\hline
\multirow{6}{*}{\textbf{Sensors}} & Encoders & 12\,bit/rev\\
& Joint torque & 12\,bit\\
& Gyroscope & 3-axis (L3G4200D chip)\\
& Accelerometer & 3-axis (LIS331DLH chip)\\
& Camera & Logitech C905 (720p)\\
& Camera lens & Wide-angle lens with 150\degree\!FOV\\
\hline
\end{tabular}
\end{table}

The \nopx builds on the ideas of providing an open-source, larger-sized humanoid robot to the research community
at a fraction of the cost of similarly sized platforms. At a height of \SI{135}{cm} and approximately \SI{19}{kg} of weight,
it aims to provide a powerful middle-ground between smaller low-cost robots, and larger cost-prohibitive platforms.
To achieve this, we utilise a combination of 3D printing, commercially available actuators, parallel kinematics and external 
gearing. Not only does this reduce cost, but also complexity and production time. The total time consumed to produce a 
working robot~(including the design stage and software adaptation) was only three months. Apart from screws, nuts, bearings, actuators, and electronics, the 
entirety of the robot is 3D printed using Selective Laser Sintering (SLS). To support the weight of the robot with inexpensive
actuators, parallel kinematics~(in the leg pitch joints) and gear transmissions~(in the leg roll and yaw joints) have been used. 
In total, the robot uses 34 actuators for 18 joints in a similar kinematic structure as the \noptwo, which was a starting point for our work.
With the \nopx, not only have we increased the capabilities, but the presented design improvements contributed to an estimate cost 
reduction of \SI{17}{\%}. Price for the \nopx parts is in a similar range to that of the popular stationary dual-arm research platform Baxter. 
A summary of the main technical specifications can be found in \tabref{OP2X_specs}.

\subsection{Intelligent Actuators}

Although the appearance of the robot may resemble that of the \noptwo~\cite{ficht2017nop2}, it is in fact
a complete redesign that incorporates several upgrades. This redesign was dictated by the choice to use a new 
series of intelligent actuators---the Robotis Dynamixel XM-540. Some of the more notable physical improvements include a \SI{29}{\%}
increase in output torque, metal casing which provides better heat dissipation and higher durability in combination
with a redesigned gearbox. What truly separates the two robots in terms of actuation, are the new 
features provided. We have modified our open-source software and firmware to take full advantage of the new hardware.
By redesigning our joint control strategy to a more abstract representation, it is now feasible to utilise the 
new possibilities, such as joint torque sensing and control. 
An added feature is the ability to effortlessly
connect sensors onto the Dynamixel bus, without having to develop additional electronics or software. Up to three analog or digital devices 
can be connected to a single actuator and operated using the standard protocol. Paired with the 3D printed nature of the design,
the platform can be adapted to many applications. 

\begin{figure}[!t]
\parbox{\linewidth}{\centering
\includegraphics[width=0.4\linewidth]{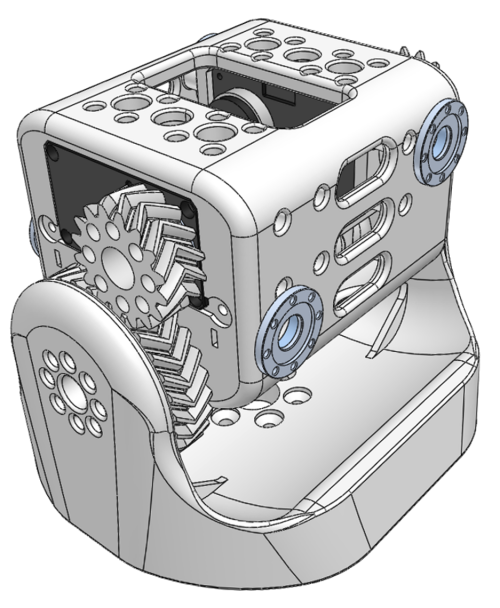}\hspace{0.1\linewidth}
\includegraphics[width=0.4\linewidth]{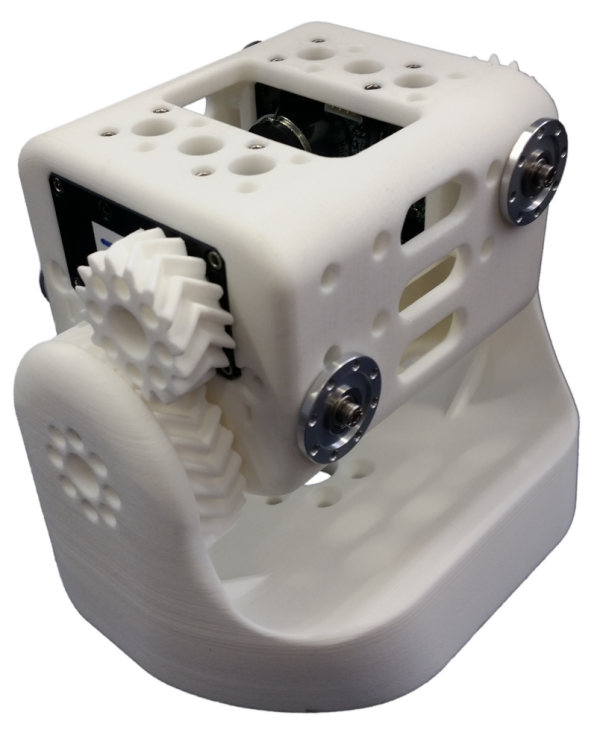}}
\caption{Redesigned hip joint utilising 3D printed double helical gears. CAD model (left) in comparison to the real joint (right).}
\figlabel{gear_joint}
\vspace{-1ex}
\end{figure}

\subsection{Parallel Computing}

The current trend in computing is shifting towards machine learning techniques, which have been successfully applied to some of the hardest 
tasks in selected research areas with high efficiency. The most prominent examples include computer vision and 
control systems---both of which strongly relate to humanoid robotics. The \noptwo already provided substantial computing power, however
the limited space in the trunk did not allow for easy GPU integration without redesign. We have increased the available space and fitted
a Mini-ITX form-factor mainboard with a standard GPU inserted into the available PCIe slot. Combining the newest Intel Core 
i7-8700T~(6~cores with 12~threads) processor and the Nvidia GTX 1050 Ti~(768~CUDA cores) GPU allowed us to achieve significant performance at a low TDP of \SI{105}{W}.
Power is supplied to the computer through a wide-input industrial \SI{250}{W} M4-ATX on-board power supply. 
Future upgrades of the computing unit pose no problem and can be done at the component level, as every part is standardised. 
This improvement reflects our design goal of providing a highly capable and customisable platform.

\subsection{Design Optimisations}

Minimising the effort required to produce and assemble a large, customisable humanoid robot has been our goal ever since 
our collaboration with igus GmbH on their Humanoid Open Platform~\cite{Allgeuer2015b}. The \noptwo was a considerable advancement in this regard, however there were 
some issues that were left unaddressed. Despite 3D printing the whole structure of the robot in PA12 nylon, external gearing
required to support the robot weight in selected joints still had to be milled from brass. The production of these custom gears 
was costly and time consuming, which made it a bottleneck in the whole production process. With the advancement of 3D printing
materials and technology, it was possible to design gears that are in every way superior to their brass counterparts. 

We decided to use double helical gears over spur gears, due to their higher torque density and smoother operation. Subtractive manufacturing of
such gears from metal would be very expensive, as they cannot be cut with simple gear hobs. This is not a limitation when 3D printing. 
When designing a helical gear, one needs to select a helix angle $\psi$, as it is not standardised. Typical values start at 
\SI{15}{\degree} and do not exceed \SI{45}{\degree}---the most common ones being \SI{15}{}, \SI{23}{}, \SI{30}{}, and \SI{45}{\degree}~\cite{gopinath2010gears}. 
Higher values provide smoother operation at the cost of higher axial load. The helix angle also influences the gear diameter $d$---which
in standard spur gears is the product of the module $m$ and the number of teeth $z$. In helical gears, this equation is divided by $\cos\psi$.
We decided to use a $\psi$ of \SI{41.4096}{\degree}, with a module $m$ of \SI{1.5}{}. With this combination, the diameter
is always double the teeth number, as $\cos\psi = 0.75$. Not only does this provide smooth operation, but allows to design the transmission 
to be compatible with spur gears in standard modules of 1 and 2. As the two halves of a double helical gear
produce counteracting axial forces, the axial load factor is mitigated. The gears have been printed using SLS technology from
the igus I6 material, which was designed specifically for gears. Gears made from it have a low friction coefficient and do not require lubrication.
Thickness has been selected to be \SI{14}{mm}~(\SI{7}{mm} for each helical half) to increase tooth overlap. Despite increasing the thickness
almost threefold~(from \SI{5}{mm} in the \noptwo), the plastic gears are on average only \SI{54}{\%} of the brass gear weight. The achieved end-result is an elegant, smooth, light and 
inexpensive transmission that is fast to produce. An example of the gear used in the hip joint can be seen in \figref{gear_joint}.

Optimising the design was one of the greater challenges with this platform. New features provided by the actuators and 
computing unit came at a cost of increased weight. The additional \SI{1.4}{kg} meant a hypothetical increase of \SI{8}{\%} in robot weight. 
We were able to redesign the leg parts to be lighter by almost \SI{0.5}{kg}, while making them structurally stronger at the same time.
\noptwo utilised quite simple shapes and flat surfaces. This decreased weight, but made the legs prone to slight bending. 
Legs in the \nopx are rounder, slightly narrower and have dedicated cable pathways---all of which contribute to the 
overall durability and torsional resistance of the structure. Offsetting the weight increase enabled us to take full advantage 
of the torque benefits provided by the new actuators.

%% file: software.tex
\section{Software Design}
\seclabel{software_design}

\begin{figure}[!t]
\hspace{-2ex}
\parbox{\linewidth}{\centering
\includegraphics[width=0.8\linewidth]{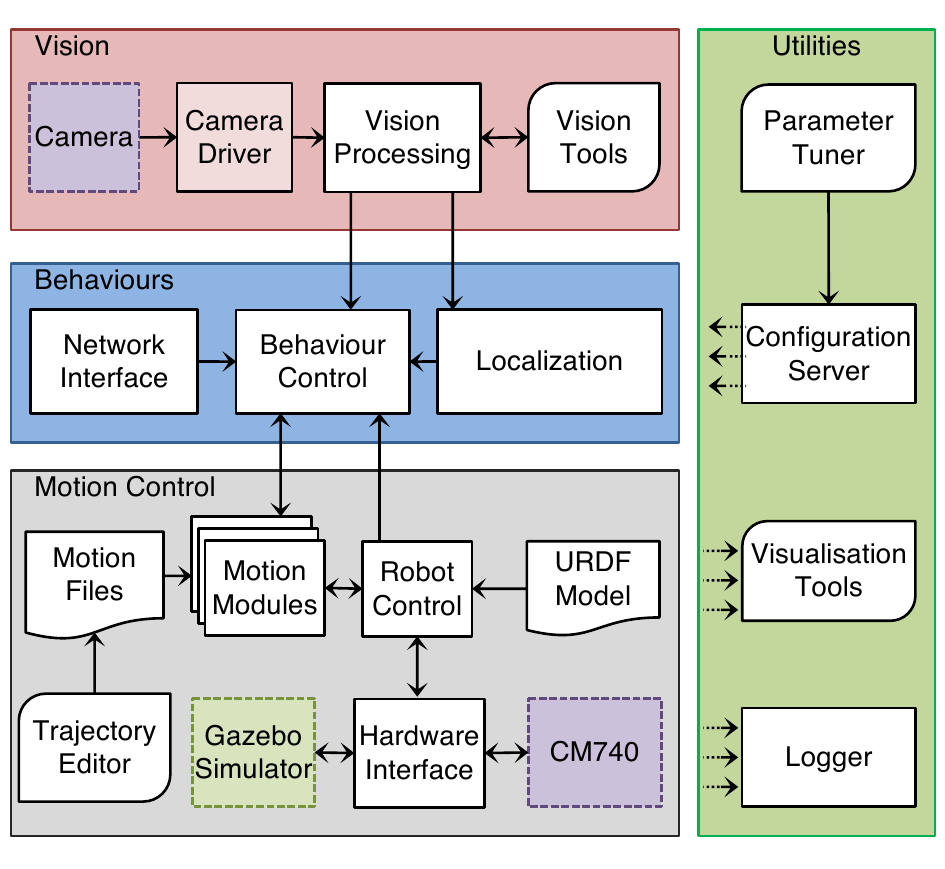}}
\caption{Architecture of our open-source ROS framework.}
\figlabel{software_architecture}
\vspace{-2ex}
\end{figure}
\begin{figure}[!t]
\hspace{-2ex}
\parbox{\linewidth}{\centering
\includegraphics[width=0.95\linewidth]{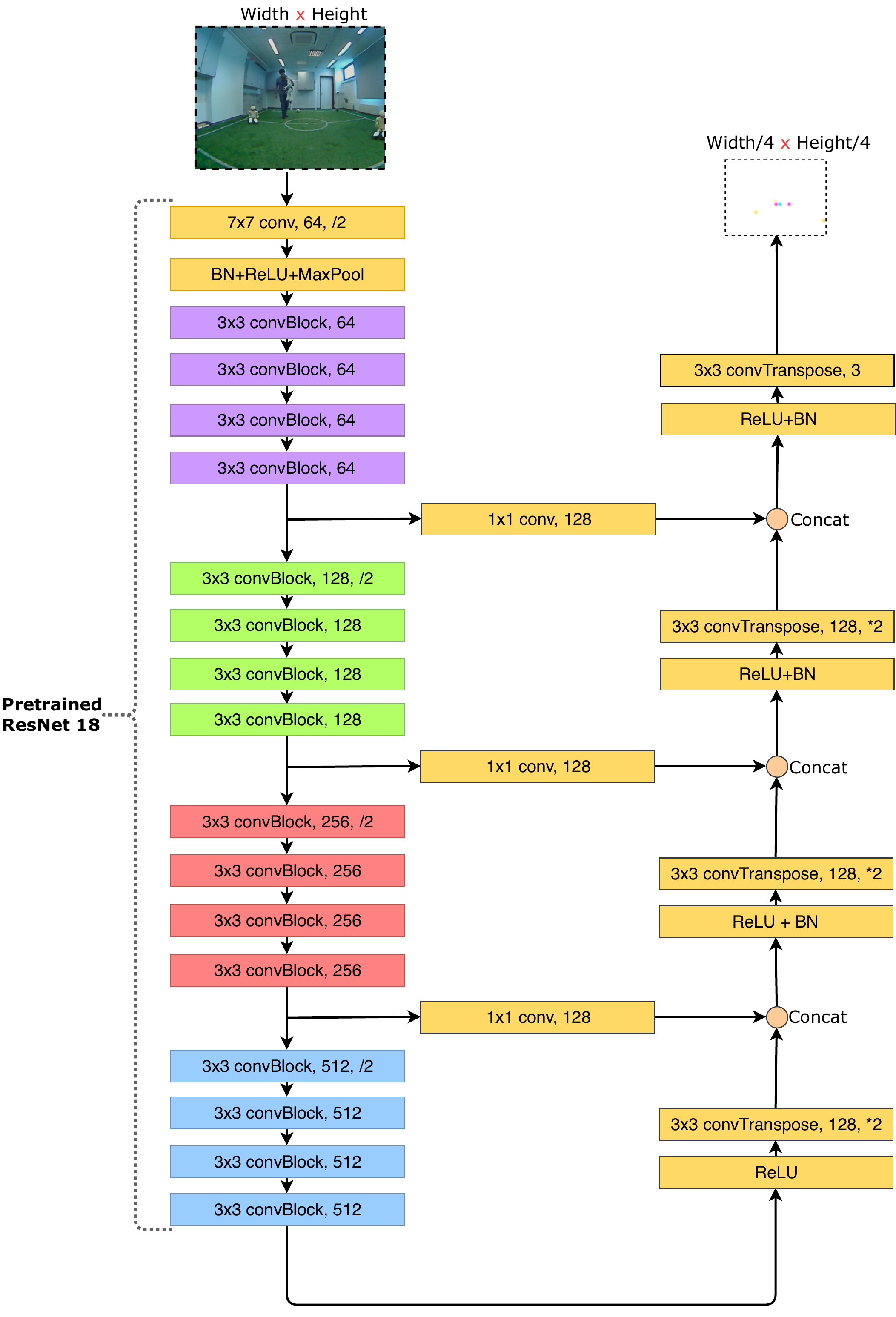}}
\caption{Visual perception network architecture. Like original ResNet architecture, each convBlock is consist of two convolutional layers followed by batch-norm and ReLU activation layers. Note that residual connections in ResNet are not depicted here.}
\figlabel{net}
\vspace{-2ex}
\end{figure}
Capable hardware is essential for reliable performance, although it is the software that truly defines the capabilities of the robot.
The software of the \nopx is heavily based on ROS~\cite{Quigley2009}. It is the result of our continued efforts
aimed at providing the research community with an easy to use, and modular framework. Five years since its creation in 2013~\cite{Schwarz2013},
it still continues to expand on functionality. Some of the more notable features include: a Gazebo simulator,
complete vision pipeline for robot soccer, tunable gait as well as a motion designer and player. A simplified schematic of the framework can be seen in \figref{software_architecture}. 
The source code is accessible online~\cite{IguhopSoftware}. With the \nopx, we focused
on developing robust vision software with deep learning methods and optimising the tunable gait parameters. Previously developed
features along with an in-depth description of the framework's inner workings can be found in~\cite{Allgeuer2013a, ficht2017nop2}.

\subsection{Visual Perception}
\nopx perceives the environment using a Logitech C905 camera which 
is equipped with a wide-angle lens and an infrared cut-off filter. 
With the introduced ability to utilise parallel computing, we are able to supersede
our previous approach to vision \cite{farazi2015}, by using a deep neural network 
in conjunction with some post-processing. We highlight the parallel computing 
capabilities of the new platform in the context of our demonstration scenario of robot 
soccer. The presented vision system can work with different viewing angles, 
brightnesses, and lens distortions. 

\begin{figure*}[!t]
\parbox{\linewidth}{\centering
\includegraphics[width=0.95\linewidth]{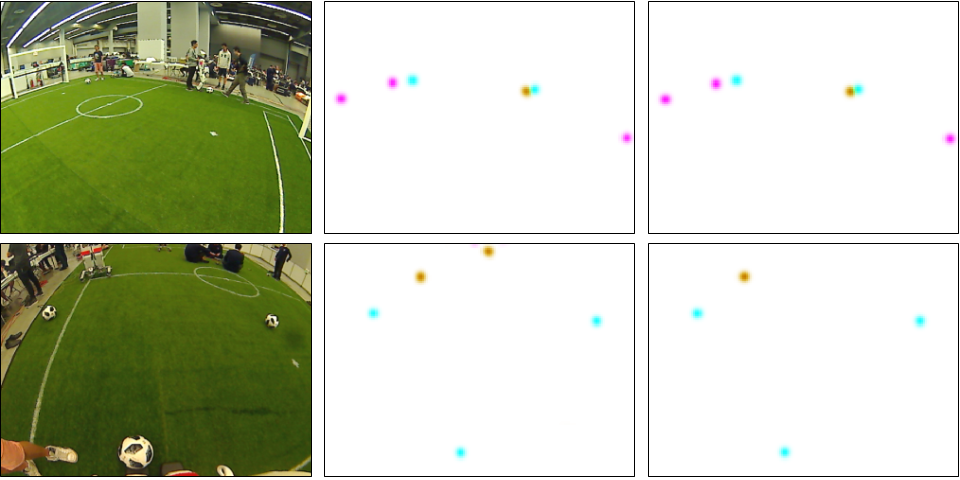}}
\caption{Object detection. Left column: A captured image from the robot in the test set. Middle column: The output of the network with balls (cyan), goal posts (magenta), and robots (yellow) annotated. Right column: Ground truth.}
\figlabel{vision_output}
\vspace{-2ex}
\end{figure*}

Our visual perception architecture is illustrated in \figref{net}. We follow the encoder-decoder approach of recently proposed segmentation 
models like SegNet \cite{badrinarayanan2015segnet}, V-Net \cite{milletari2016v}, 
and U-Net \cite{ronneberger2015u}. To achieve real-time computation and to minimize the 
number of parameters in the network, we designed the decoder part to be shorter 
than the encoder part. This design choice has the drawback that the output of the 
network has less spatial information than the full resolution input. We mitigate 
this effect by using a subpixel centroid-finding method in the post-processing 
step. In order to minimize the training time, we lower the number of samples in the 
dataset, and leverage big datasets like ImageNet, we used a pre-trained ResNet-18 
model for our encoder part. As shown in \figref{net}, we removed the last fully 
connected and GAP layers in the ResNet-18 model. Similar to the U-Net model, we also have lateral 
connections between encoder and decoder parts to directly provide high-resolution details 
to the decoder. In total, we have 23 convolutional layers 
and four convolution-transpose layers. 

The network is trained on the following object classes: ball, goal post, and 
robots. Instead of using the full segmentation loss, similar to SweatyNet \cite{schnekenburger2017detection}, 
we use a Gaussian blob around the center of the ball and the bottom-middle 
point of the goal post and robots. As center point detection is enough for objects mentioned earlier, instead of full semantic segmentation loss, mean squared error is used as the loss function.
To find a good learning rate, we followed the approach presented by Smith 
\cite{smith2017cyclical}. We have also tested the recently proposed AMSGrad optimizer 
\cite{2018on}, but did not see any benefit in our application, over the employed Adam.
To train the network as fast as possible, we first downsample the training 
images and train for 50 epochs while freezing the weights on the encoder part. Freezing the encoder part during the first epochs quickly adjusts the randomly initialized weights in the decoder part, that is inspired by the work of Brock et al.~\cite{brock2017freezeout} and by the paper from Yosinski et al.~\cite{yosinski2014transferable}. During the next
50 epochs, all parts of the models rae trained. Finally, full-sized images
are used to train the model for 50 more epochs. For the pre-trained model, a lower learning
rate is used. In total, the whole training process with around 3000 samples takes less than 40 
minutes on a single Titan Black GPU with 6\,GB memory.  Some of the training 
samples were taken from the ImageTagger library \cite{imagetagger2018} which 
have annotated samples from different angles, cameras, and brightness. 
Although the network produced very few false-positives (around \SI{1}{\%} for the ball), 
we were able to reduce this value by utilising inference time augmentation.
Two samples of the testset are depicted in \figref{vision_output}.
Extracting object coordinates is done with minor post-processing, since the output of the network is blob-shaped.
After thresholding each output channel, we apply morphological erosion and dilation to eliminate negligible 
responses. Finally, we compute the object center coordinates from the achieved contours.

After detecting game-related objects, we filter them and project each object location into egocentric world coordinates. 
These coordinates then are further processed in the behaviour node of our ROS-based 
open-source software for decision making. To alleviate projection errors due to the 
slight differences between the CAD model and real hardware, we calibrate the camera 
frame offsets, using the Nelder-Mead~\cite{nelder1965simplex} Simplex method.

Through the improvements made on our vision system, we were able to perceive a FIFA size 5 ball
up to \SI{7}{m} with an accuracy of 99\% and less than 1\% of false detections. Goal 
posts can be detected up to 8 meters with 98\% precision and with 3\% false 
detection rate. Robots are detected up to 7 meters with a success rate of 90\% 
and a false detection rate of 8\%. For line and field detections, we are using non-deep learning approaches \cite{farazi2015}, 
but are working on adding two more channels to the network output and use a single network 
for all detections. The complete vision pipeline including a forward-pass of the network takes 
approximately 20\,ms on the robots hardware.

\input{bayesopt.tex}

%% file: bayesopt.tex
\subsection{Sample-efficient Gait Optimisation}

Perceiving the environment is only the first step to interacting with it. An essential skill for 
humanoid robots is walking. The gait of the \nopx is based on an 
open-loop pattern generator that calculates joint states based on a gait phase angle, 
whose rate is proportional to the desired step frequency~\cite{Behnke2006}.
The phase angle is responsible for generating arm and leg movements such as lifting and swinging.
We have built around this approach and incorporated corrective actions based on fused angle feedback~\cite{Missura2013a, Allgeuer2015, Allgeuer2016a}.

One of the drawbacks of this method, is having to tune the gait for each and every robot, as they 
can differ not only in structure but have slight mechanical differences between theoretically identical robots.
We remedy this by using optimisation to find proper values for the Fused Feedback 
controller used by the gait. To minimise hardware wear-off, this optimisation does 
not only take place in the real world, but highly exploits information gained 
through the included Gazebo simulator. This approach has been previously applied 
by Rodriguez et al.~\cite{Rodriguez2018Combining} and is now utilised on the \nopx robot. 

\begin{figure}[!t]
  \footnotesize
  \centering
  
  \includegraphics[width=\linewidth]{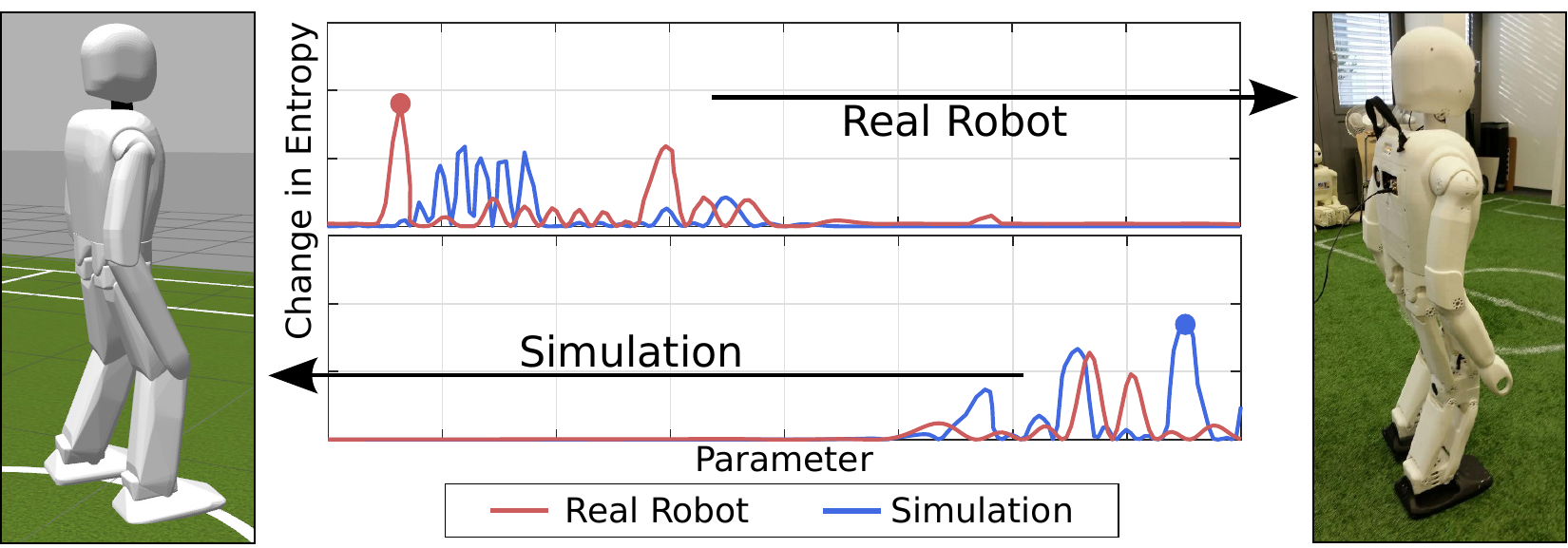}
  \caption[]{At each iteration the algorithm trades off performing a simulation 
or a real-world experiment. This decision is led by the expected change of 
Entropy, making the algorithm particularly sample-efficient.}
\label{fig:bayesteaser}
\end{figure}

Since we pursue the goal of optimising a given parameter set in a 
sample-efficient manner, the proposed method uses \textit{Bayesian 
Optimisation}. This allows to efficiently trade off exploration and 
exploitation. It requires to specify a 
\textit{Kernel function}~$K$ and a parametrisation of a \textit{Gaussian 
Process}~(GP). Both of these are responsible for encoding essential information about the optimisation environment.
We use a kernel that is comprised of two terms. The first one relates to the simulation performance $k_{sim}$, while the other is an
error term $k_\epsilon$ depicting the difference between simulations and the 
real world. This results in: 
\begin{equation}\label{eq:compositeKernel} 
k(\mathbf{a_i},\mathbf{a_j})=k_{sim}(\mathbf{x_i},\mathbf{x_j} 
)+k_\delta(\delta_i,\delta_j)k_{\epsilon}(\mathbf{x_i}, \mathbf{x_j})\, 
\end{equation} 
for an augmented parameter vector $\mathbf{a_i}=(\mathbf{x_i},\delta_i)$, where 
$\delta$ signals whether the evaluation has been performed in simulation or 
on the real system. An indication if both experiments were in the 
real world is $k_\delta$, which returns a higher correlation in that case. This way, it is 
possible to address potentially complex, non-linear transformations between the 
simulated sequences and the real-world performance. To perform the optimisation 
process in a data-efficient manner, the approach utilises \textit{Entropy} as a way of
measuring the information content. The next evaluation point is chosen with respect to
the maximal achievable change in Entropy~\cite{hennig2012}, biased by 
weight factors trading off simulation and real-robot experiments (see Fig. 
\ref{fig:bayesteaser}). 

For both kernels, we select the Rational-Quadratic Kernel, since it has proven to 
be suitable in our previous work~\cite{Rodriguez2018Combining}. 

The proposed cost function is based on the proportional Fused 
Feedback $e_{P}$, which employs the Fused Angle deviations and thus can be 
considered a stability measure~\cite{Rodriguez2018Combining}. Furthermore, to 
reduce the impact of noise, we examine the cost of the sagittal($\alpha$) and 
lateral($\beta$) planes separately. Since higher gain values lead to higher 
torques, we also penalise the magnitude of the parameter using a logistic 
function $\nu$. This results in the cost functions:
\begin{equation}\label{eq:finala}
J_{\alpha}(\mathbf{x}) = \int_0^T{\Vert e_{P\alpha}(\mathbf{x})\Vert_1}dt + 
\nu(\mathbf{x})
\end{equation}
\begin{equation}\label{eq:finalb}
J_{\beta}(\mathbf{x}) = \int_0^T{\Vert e_{P\beta}(\mathbf{x})\Vert_1}dt + 
\nu(\mathbf{x})\,,
\end{equation}
where $e_P$ is seen as a function of the used gait parameters $\mathbf{x}$. 

\begin{figure}[!b]
	\footnotesize
	\centering

	  \resizebox{\width}{0.92\height}{\includegraphics[width=0.8\linewidth]{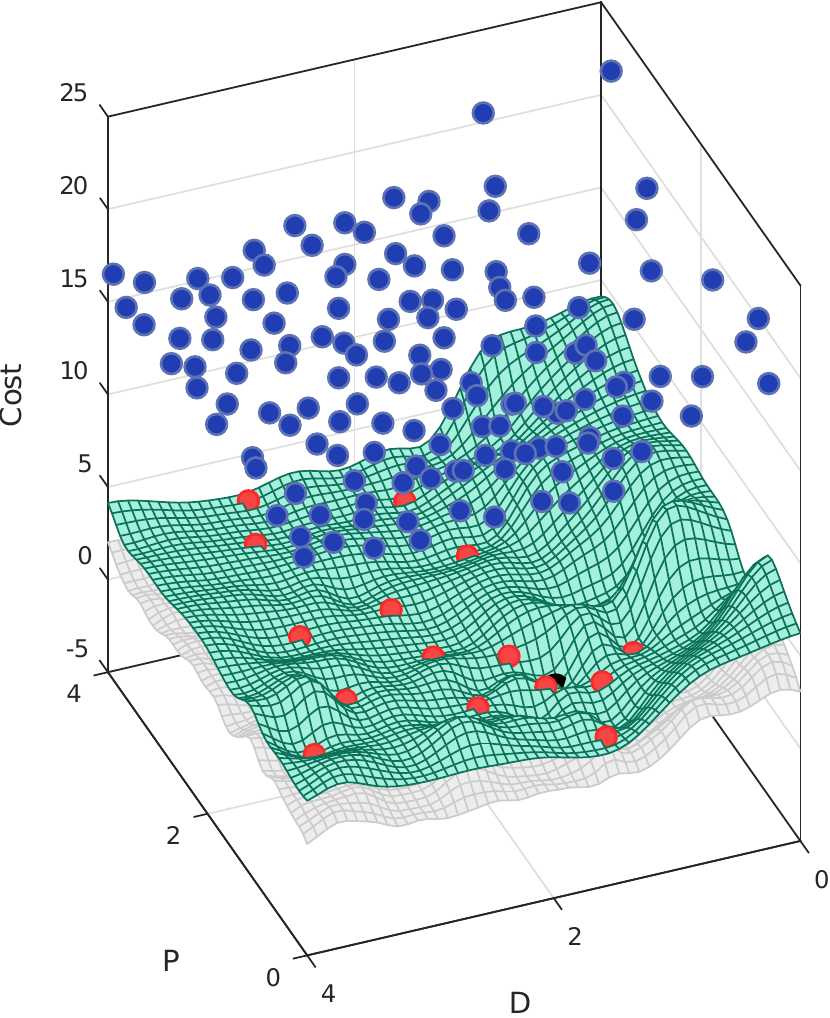}}
	\caption[]{The GP posterior of the Arm Angle optimisation is 
depicted. The blue dots resemble evaluations in simulation, while real 
experiments are shown in red. A black dot indicates the estimated minimum. The 
mean cost estimate is displayed (teal mesh), together with its standard 
deviation (grey mesh). To increase visibility, the upper standard deviation is 
not shown.}
 \label{fig:gppost}
\end{figure}

An evaluation of the cost function is done through a complex test sequence consisting of the same, pre-defined 
combination of forward, backward and sideways gait commands. %
To receive more reproducible results in simulations, we average the cost of $N=4$ simulated runs. 
In the described example, we optimise the Arm Angle corrective actions in the sagittal 
direction. More precisely, we restrict our search to the P and D gains.
The method can be similarly applied to other controllers. To 
preserve the robot's hardware, we restrict the permitted number of real-world 
evaluations to 15, which is comparable to the time it takes an expert to 
hand-tune the parameters. These 15 runs were accomplished after a total of 161 
iterations, thus including 146 simulated gait sequence evaluations. To assess 
the resulting parameters, the deviations of the new parameters were compared to 
the old parameters in 5 evaluation sequences each. The results of the optimisation not only
produces a qualitatively more convincing walk, but also leads to a reduction of the 
Fused Angle deviation by approximately \SI{18}{\%}.

The GP posterior of the optimisation is depicted in Fig.~\ref{fig:gppost}. Although 
it might seem that the simulations only had a small influence on the result, 
they were of utmost importance in early iterations~\cite{Rodriguez2018Combining}. 
This is underlined especially by the fact, that the robot did not fall a single 
time during real-world evaluations, which proves that the GP model is able to 
utilise simulation data to rule out inferior parameters. This stems from the high
costs associated with unstable simulated runs. 

Altogether, our approach has proven to be beneficial with the \nopx, and 
yielded successful parameterisations for the used gait. Thus, it allows for 
easy operation even for an inexperienced user. It can be a feasible alternative 
to the time-consuming hand-tuning performed by an expert, which wears off the 
robot's hardware.

%% file: evaluation.tex
\begin{figure*}[t]
\centering
\includegraphics[height=0.27\linewidth]{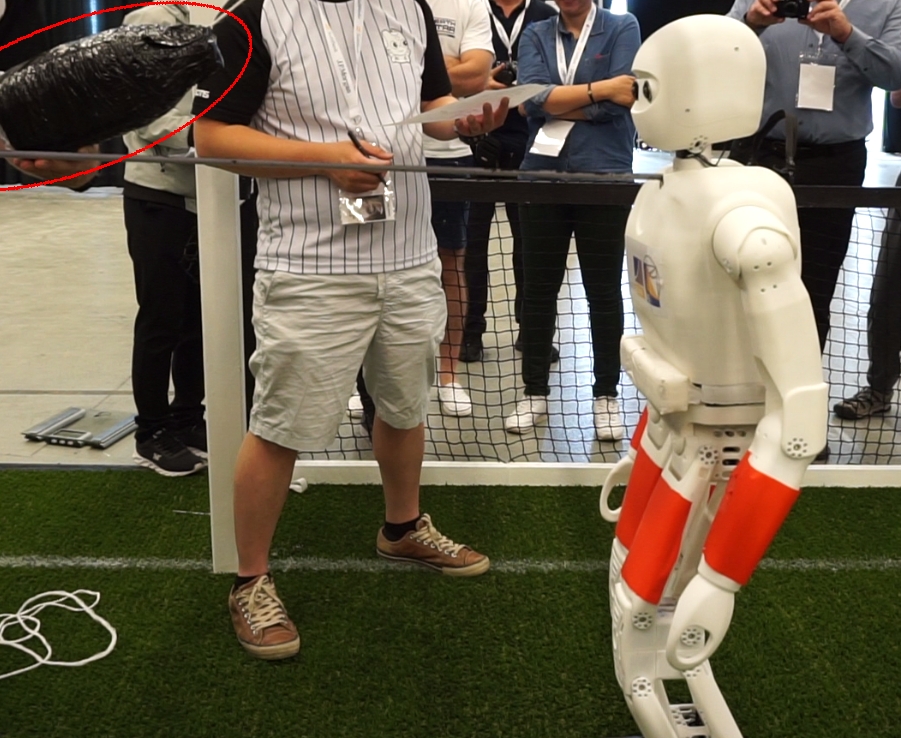}\hspace{1px}
\includegraphics[height=0.27\linewidth]{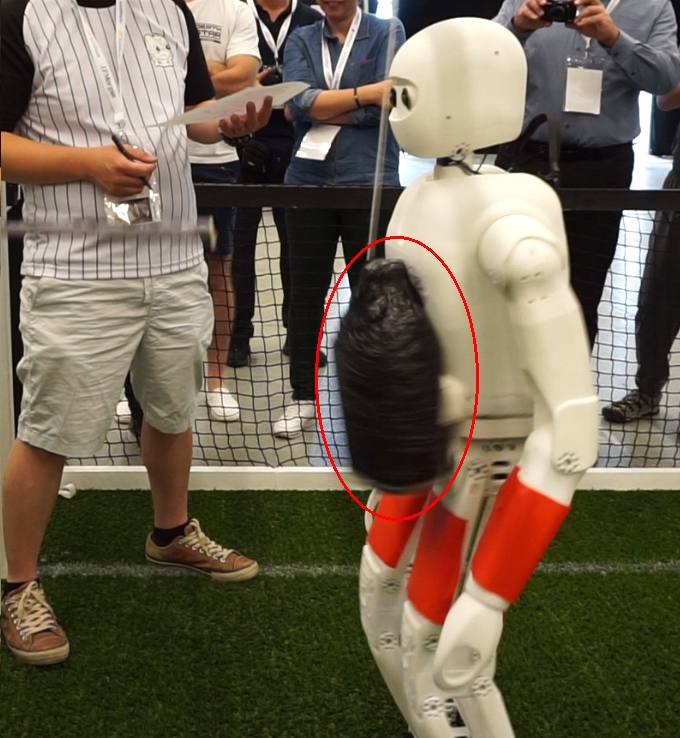}\hspace{1px}
\includegraphics[height=0.27\linewidth]{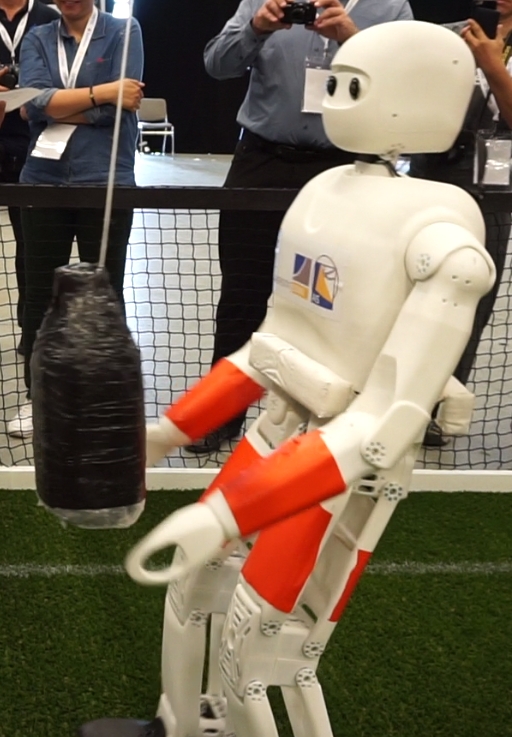}\hspace{1px}
\includegraphics[height=0.27\linewidth]{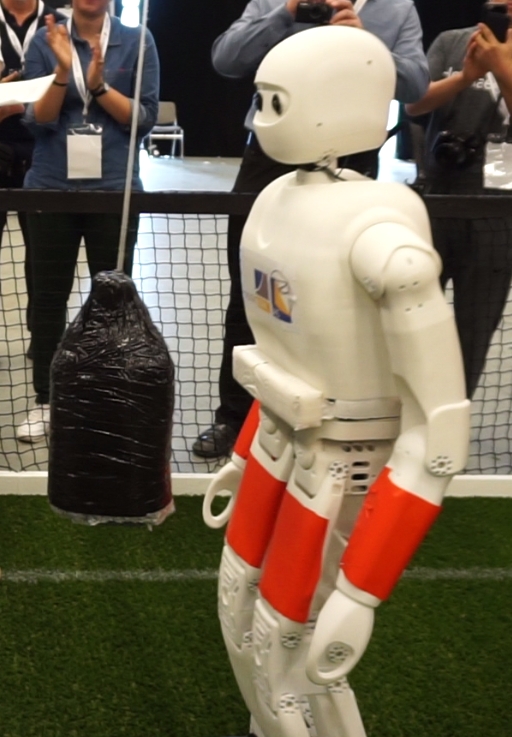}\vspace*{-1ex}
  \caption{\nopx withstanding a push from the front. The weight is marked in red. After receiving the push, the robot performs corrective actions to recover stability.}
  \figlabel{push_recovery}\vspace{-2ex}
\end{figure*}
\section{Evaluation}

An evaluation of the \nopx was performed during the RoboCup 2018 competition in Montr\'eal, Canada.
In the RoboCup Humanoid League AdultSize class, robots autonomously compete in one vs. one soccer games, two vs. two drop-in games, and four technical challenges that test specific abilities in isolation.
The soccer games are performed on a $6\times9$\,\SI{}{m} artificial grass field, which makes walking challenging.
Perceiving the environment and localisation are also challenging, due to the ever changing lighting conditions.
In the competition, the \nopx performed outstandingly by winning all out of the four possible awards, including the Best Humanoid Award.  %

In the main tournament, our robot played a total of six games, including the quarter-finals, semi-finals and finals (see \figref{soccer}).
An additional five drop-in games were played, where two vs. two mixed teams were formed and collaborated during the game.
This resulted in 220 minutes of official play time, during which 66 goals were scored and only 5 were conceded.

The optimised gait was not only fast~(approx. \SI{0.5}{m/s}) but also robust, as \nopx never fell during 
walking in free space or while dribbling the ball. Very strong collisions with much heavier robots made the robot fall over four times.

One of the biggest contributions to the success, was our vision system utilising deep learning.
The ball detections were robust up to a distance of \SI{7}{m}, where the perceived ball was only a few pixels in size.
Robot detections allowed for efficient fights for the ball, which often resulted in the ball being dribbled around the opponent.
This was a decisive factor during the final, where the \nopx won 2:0 against Sweaty from Hochschule Offenburg, Germany.
The opponent played very defensively and was efficient at dribbling and handling the ball.
\begin{figure}[b]
\centering
\includegraphics[width=0.99\linewidth]{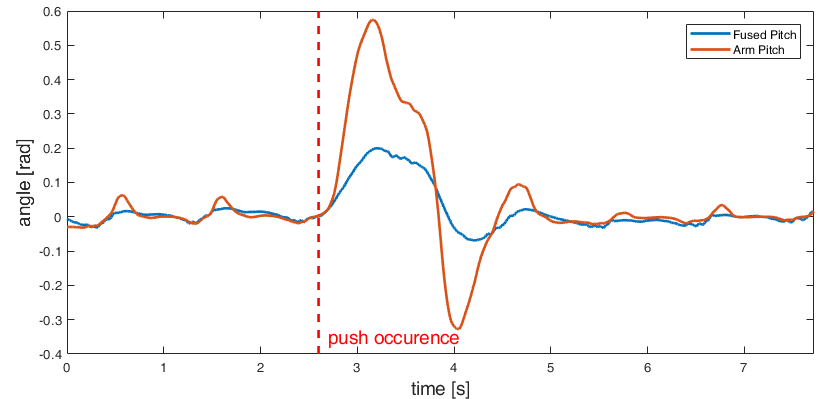}\vspace*{-1ex}
  \caption{Push Recovery trial at RoboCup 2018. Optimised corrective arm action in response to expected angle deviation in the sagittal plane.}
  \figlabel{push_graph}\vspace{-2ex}
\end{figure}

Performance of the vision and gait was also tested during the technical challenges.
In the Push Recovery challenge, the goal is to withstand pushes from the back and from the front.
The disturbance is applied by a \SI{3}{kg} heavy pendulum swinging on a \SI{1.5}{m} long string.
The weight hits the robot at the height of the center of mass. 
In order to perform the push, the weight is retracted from the robot by a distance $d$ and then released freely.
\nopx was able to successfully withstand both frontal and backward pushes with $d = 0.8$\,m.
The optimised arm swing corrective action helped substantially---its effect can be seen in \figref{push_recovery} and \figref{push_graph}.
A different challenge required the robot to score a goal from a moving ball which is passed by a 
human player from the corner of the field. Our fast image processing and capable hardware allowed 
the robot to score a goal in approximately 3 seconds, counting from when the ball was passed.

\begin{figure}[b]
\centering
\includegraphics[height=0.73\linewidth]{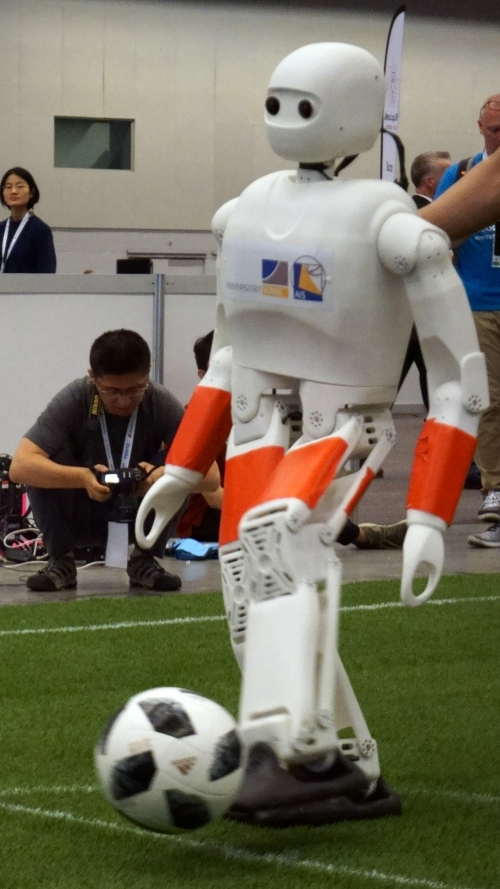}\hspace{1px}
\includegraphics[height=0.73\linewidth]{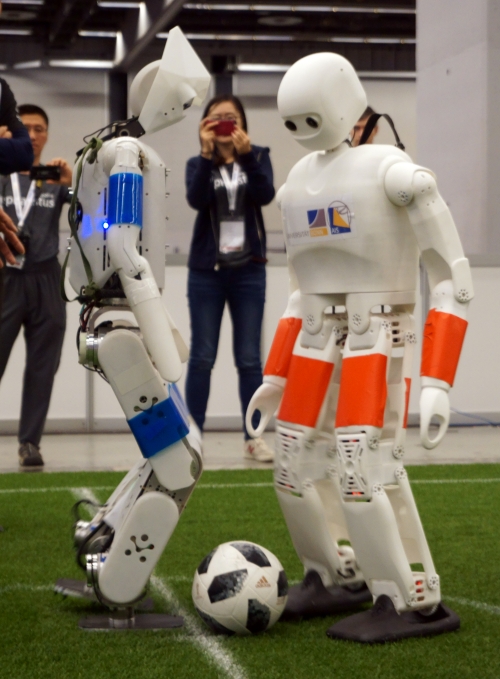}\vspace*{-1ex}
  \caption{\nopx during RoboCup 2018. Left: Performing a kick. Right: Competing for the ball.}
  \figlabel{soccer}\vspace{-2ex}
\end{figure}

%% file: conclusions.tex
\section{Conclusions}

In this paper, we have described the newly developed hardware and software aspects of the \nopx. 
The technical specifications and provided features make it a powerful middle-ground for the research community
between smaller, inexpensive robots and larger cost-prohibitive platforms.
Not only is \nopx large enough to interact with a human environment, its relatively light weight
makes it easy and safe to operate. The minimalistic 3D printed design allows for easy maintenance as well as user 
modifications and improvements. By demonstrating efficient use of its hardware
and new deep-learning capabilities, the robot was able to outperform its competition at RoboCup 2018. 
We plan on releasing the 3D-printable hardware components with a Bill of Materials, as well as updating our open-source software~\cite{IguhopSoftware} by October 2018.